\pdfoutput=1
\documentclass[final]{cvpr}

\usepackage{times}
\usepackage{graphicx}
\usepackage{amsmath}
\usepackage{amssymb}
\usepackage{subfigure}
\usepackage{mmstyles}

\usepackage[pagebackref=true,breaklinks=true,colorlinks,bookmarks=false]{hyperref}

\newcommand{\method}{\textcolor{black}{BRNet }}
\newcommand{\methodnospace}{\textcolor{black}{BRNet}} 

\def\cvprPaperID{2149} 
\def\confYear{CVPR 2021}

\begin{document}

\title{
Back-tracing Representative Points for Voting-based \\
3D Object Detection in Point Clouds
}

\author{
    Bowen Cheng$^1$,~~
    Lu Sheng$^1$\thanks{Lu Sheng is the corresponding author.},~~
    Shaoshuai Shi$^2$,~~
    Ming Yang$^1$,~~
    Dong Xu$^3$ \\
    $^1$College of Software, Beihang University \\ $^2$The Chinese University of Hong Kong ~~ $^3$The University of Sydney \\
    \small
    \texttt{\{chengbowen052, lsheng, viv\}@buaa.edu.cn} \hspace{20pt}
    \texttt{ssshi@ee.cuhk.edu.hk} \hspace{20pt}
    \texttt{dong.xu@sydney.edu.au}
}

\maketitle


\begin{abstract}
    3D object detection in point clouds is a challenging vision task that benefits various applications for understanding the 3D visual world.
    Lots of recent research focuses on how to exploit end-to-end trainable Hough voting for generating object proposals.
    However, the current voting strategy can only receive partial votes from the surfaces of potential objects together with severe outlier votes from the cluttered backgrounds, which hampers full utilization of the information from the input point clouds.
    Inspired by the back-tracing strategy in the conventional Hough voting methods, in this work, we introduce a new 3D object detection method, named as Back-tracing Representative Points Network (\methodnospace), which generatively back-traces the representative points from the vote centers and also revisits complementary seed points around these generated points, so as to better capture the fine local structural features surrounding the potential objects from the raw point clouds.
    Therefore, this bottom-up and then top-down strategy in our \method enforces mutual consistency between the predicted vote centers and the raw surface points and thus achieves more reliable and flexible object localization and class prediction results.
    Our \method is simple but effective, which significantly outperforms the state-of-the-art methods on two large-scale point cloud datasets, ScanNet V2 (+7.5\% in terms of mAP@0.50) and SUN RGB-D (+4.7\% in terms of mAP@0.50), while it is still lightweight and efficient.
    Code will be available at \href{https://github.com/cheng052/BRNet}{https://github.com/cheng052/BRNet}.
\end{abstract}

\section{Introduction}

As one of the fundamental tasks that aims at understanding 3D visual world, 3D object detection would like to predict amodal 3D bounding boxes and associated semantic labels of objects in real 3D scenes.
3D object detection technologies would significantly benefit various downstream real world applications such as augmented reality, robotics and \etc.
In this work, we focus on 3D object detection from point clouds.
It is even more challenging because the irregular, sparse and orderless characteristics of this special 3D input make it a hard task to design reliable point-based 3D object detection systems by leveraging the recent progress in 2D object detection.

While earlier works resorted to reordering point clouds into regular forms~\cite{chen2017multi,3d-sis,song2014sliding,dss,zhou2018voxelnet}, or applying predefined shape templates~\cite{li2015database,nan2012search,gspn}, VoteNet~\cite{votenet} and its variants~\cite{mlcvnet,h3dnet,hgnet, ahmed2020density} have shown a great success in designing end-to-end 3D object detection networks based on raw point clouds.
VoteNet reformulates the traditional Hough voting process into a point-wise regression problem, and generates an object proposal by sampling a number of seed points from the input point cloud whose votes are within the same cluster.
The aggregated feature in each vote cluster is then used to estimate the 3D bounding box (\eg center, size and orientation) and the associated semantic label.

\begin{figure}[t]
    \centering
    \includegraphics[width=\linewidth]{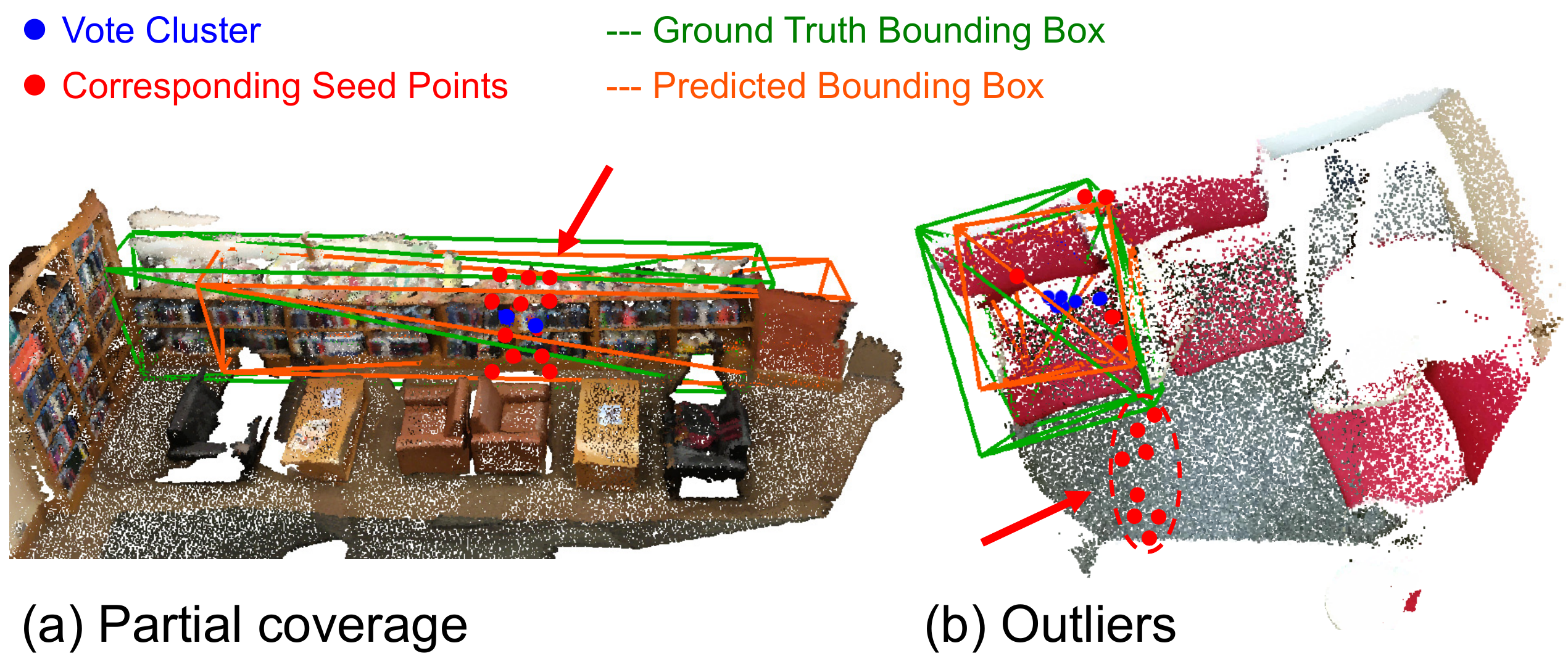}
    \caption{The votes generated by VoteNet~\cite{votenet} and its variants usually suffer from (a) partial coverage of the object surfaces, (b) outliers from the cluttered background. By examining the corresponding seed points, the generated proposals from these votes receive erratic features with respect to the objects, and may be less reliable for predicting accurate bounding boxes, orientations and even semantic classes. Best viewed on screen.}
    \label{fig:intro}
\end{figure}

Therefore, the quality of the regressed votes principally determine the reliability of the generated proposals, and then the performance on the object detector.
However, although the clustered vote centers are quite accurate, the votes are usually not as representative as our expectation.
For example, as illustrated in Fig.~\ref{fig:intro}, by retrieving the seed points of votes from the given vote clusters, these corresponding seed points either partially cover the underlying objects (Fig.~\ref{fig:intro}(a)) or contain severe outliers from the cluttered background (Fig.~\ref{fig:intro}(b)).
Therefore as shown in Fig.~\ref{fig:intro}(a), it is undoubted that we cannot accurately predict the bounding box of a long bookshelf if the votes only capture a small area surrounding the vote center.
Likewise as shown in Fig.~\ref{fig:intro}(b), the severe outliers make it impossible to accurately detect the chair based on the vote features.
Moreover, these seed points are less informative due to the lack of knowledge from the votes, so that there will be less significant gains if we simply back-trace these seed features (as in conventional Hough voting~\cite{robust-hough}) to improve the voting-based 3D object detection methods.

\begin{figure}
    \centering
    \includegraphics[width=0.7\linewidth]{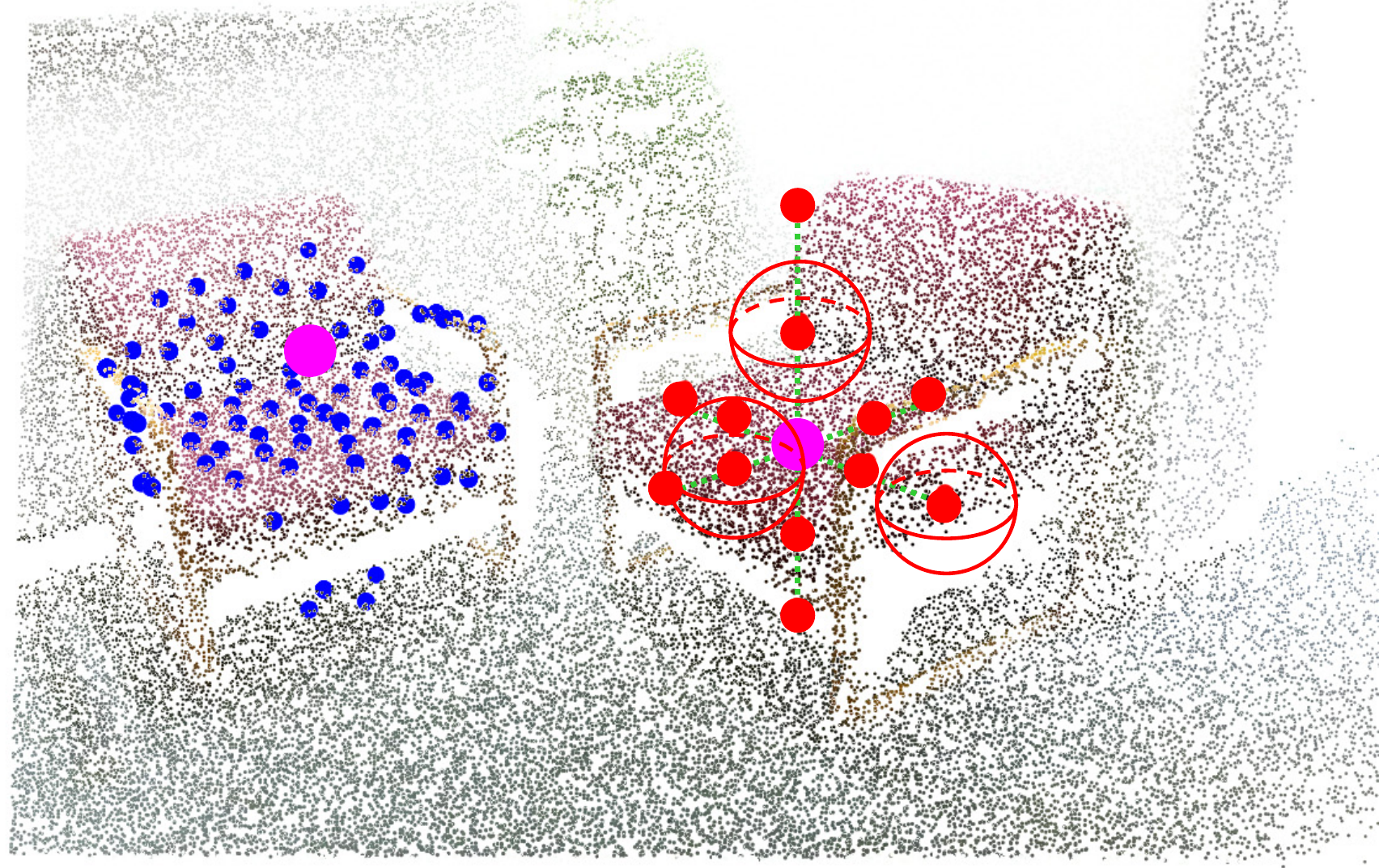}
    \caption{
        Back-tracing representative points and revisiting seed points. 
        We show the vote cluster center for the two chairs (in purple points). 
        The representative points are back-traced from the vote cluster center (in red points).
        We set the number of representative points per proposal as 12 in this case, which are illustrated on the right chair. 
        Then, the seed points within a fixed distance of the representative points are revisited, shown in blue points on the left chair. 
        The revisited seed points provide good coverage of the chair's surface, which imply the object shape and keep the structural details as the chair armrest. Best viewed on screen.}
\label{fig:illustrate-back-tracing}
\end{figure}

However, in our point of view, back-tracing is still necessary and could partially address the aforementioned issues with a special design.
To be specific, as shown in Fig.~\ref{fig:illustrate-back-tracing}, we would like to backwardly generate (or trace) the virtual representative points from the center of each vote cluster, and use these virtual points to revisit their surrounding seed points.
This generative back-tracing operation indicates possible object shape distributions around the vote center, while the revisited seed features provide complementary local structural clues that may not be fully discovered by the votes.
This bottom-up and then top-down process can end up with a mutual interaction that associates the seed features and the vote features, which has the potential to enhance each other features and enable more robust object class prediction and more accurate bounding box regression.

To this end, we propose a new point cloud-based 3D object detection method, named as Back-tracing Representative Point Network (\methodnospace), by incorporating the end-to-end learnable back-tracing and revisiting operations into the voting-based framework.
Specifically, we propose a representative points generation module that generatively samples uniformly distributed representative points within the 3D area of a candidate object, based on the features of a vote cluster center.
The generated points can coarsely infer the object bounding boxes even though their sampling process is class-agnostic.
The revisited seed points of each representative point are aggregated in a similar way as ROI grid pooling~\cite{pvrcnn}, but based on the spatial layout of the representative points.
After fusing the aggregated features of the revisited seed points and the features of the vote cluster center, we obtain the refined proposals to eventually detect the objects.
Note that the proposed bounding box regression scheme explicitly depends on the spatial distribution of the representative points, thus improves robustness with respect to shape variations within and across object categories.

The contributions of this work are three-fold:
(1) the first 3D object detection network, named as \methodnospace, that successfully adapts the back-tracing step of Hough voting to 3D object detection.
(2) an end-to-end learnable network that can generatively \emph{back-trace} the representative points, reliably \emph{revisit} the seed points, and then mutually \emph{refine} the object proposals for more robust object classification and more accurate bounding box regression.
(3) the state-of-the-art 3D object detection performance on two benchmark datasets, ScanNet V2~\cite{scannet} (50.9\% in terms of mAP@0.50) and the SUN RGB-D~\cite{sunrgb-d} (43.7\% in terms of mAP@0.50).

\section{Related Works}
\label{sec:related_works}

\noindent\textbf{3D object detection on point clouds.}
Object detection from 3D point clouds is challenging due to the irregular, sparse and orderless characteristics of 3D points.
Earlier attempts usually relied on projections onto regular grids such as multi-view images~\cite{chen2017multi} and voxel grids~\cite{zhou2018voxelnet, yan2018second, lang2019pointpillars, 3d-sis, partA2}, or based on the candidates from RGB-driven 2D proposal generation~\cite{f-pointnet,2d-driven} or segmentation hypotheses~\cite{kim2013accurate}, where the existing 2D object detection or segmentation methods based on regular image coordinates can be effortlessly adapted.
Other approaches also studied how to exploit discriminative~\cite{li2015database,nan2012search} or generative shape templates~\cite{gspn}, and high-order contextual potentials to regularize the proposal objectness~\cite{lin2013holistic}, or used sliding shapes~\cite{dss,song2014sliding}, or clouds of oriented gradients (COG)~\cite{cog}.

Thanks to PointNet~\cite{pointnet}, deep neural networks have become extensively employed onto raw point clouds.
For instance, PointRCNN~\cite{pointrcnn} introduced a two-stage 3D object detector, which is analogous to the two-stage 2D object detection methods such as Faster RCNN~\cite{faster-rcnn}.
Inspired by the Hough voting strategy for 2D object detection and instance segmentation~\cite{robust-hough}, VoteNet~\cite{votenet} was built upon the backbone of PointNet++~\cite{qi2017pointnet++} and presented an end-to-end trainable 3D object detector.
Later on, the extensions of VoteNet~\cite{votenet}, such as MLCVNet~\cite{mlcvnet}, HGNet~\cite{hgnet} and 3DSSD~\cite{yang20203dssd}, employed the contextual clues, the hierarchical graph neural networks and the feature-FPS sampling strategy to enable better generation of object proposals.
However, these methods heavily depend on the unreliable vote clustering proposed in~\cite{votenet}, which is inevitably affected by outliers and usually overlooks inlier seed points.
H3DNet~\cite{h3dnet} partially tackled this issue by introducing a hybrid set of overcomplete geometric primitives to refine the initial bounding boxes predicted by the clustered votes.
But these primitives centers are learned with less accurate supervisions and also collected by a similar clustering strategy, thus may still fail to eliminate the outliers or capture sufficient geometric clues to infer the target objects.
In this work, we show how to leverage the representative points back-traced from the vote centers to complementarily profile the target objects, which enables more discriminative categorization and more robust bounding box regression.

\begin{figure*}
    \centering
    \includegraphics[width=\linewidth]{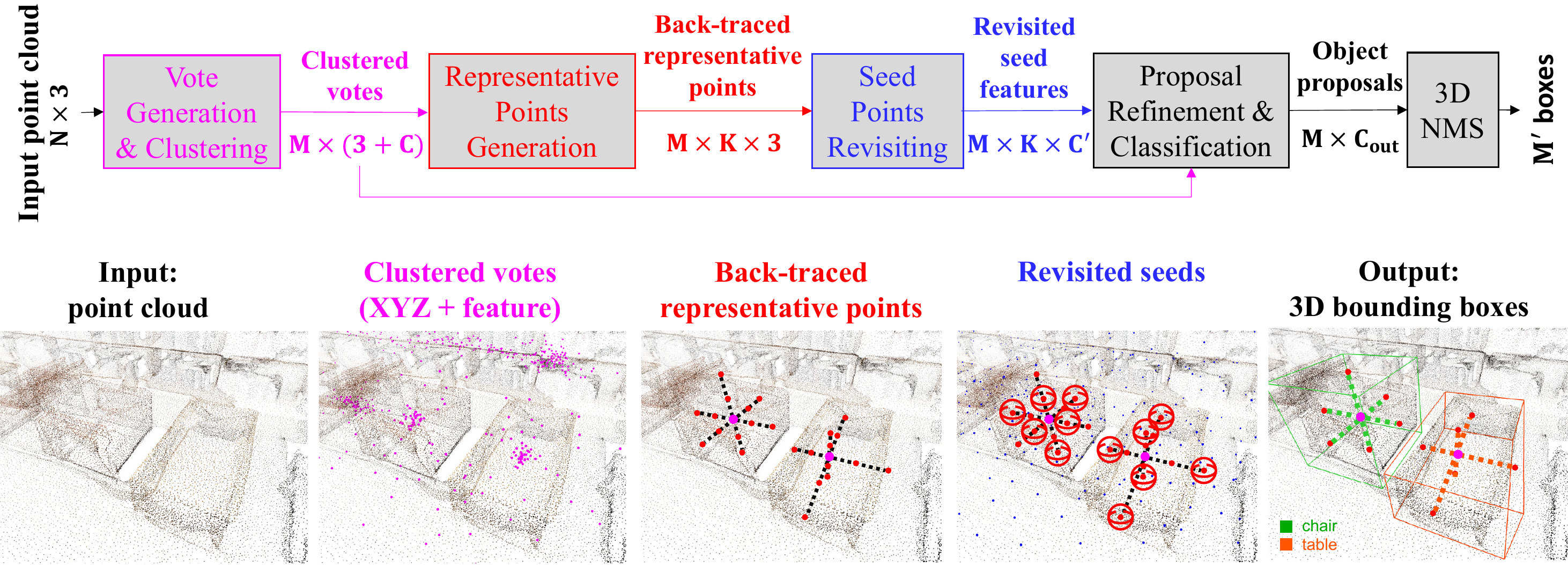}
    \caption{
    An overview of our proposed \method for 3D object detection in point clouds. 
    Given an input point cloud consisting of $N$ points with the XYZ coordinates, we generate votes from it and group the votes into $M$ clusters as in VoteNet~\cite{votenet}. 
    For each of the $M$ vote cluster centers, we back-trace $K$ representative points from it. 
    The back-traced representative points imply the possible area of the object. We then revisit the seed points around the representative points and aggregate the surrounding seed point features to the representative points. 
    The clustered vote features and the revisited seed features are fused and processed by the proposal refinement and classification module to produce the refined representative points and object's semantic category, which can be easily transformed into 3D object bounding boxes. 
    The standard 3D NMS is eventually used to generate the final detection results. Best viewed on screen.
    }
    \label{fig:method}
\end{figure*}

\vspace{+1mm}
\noindent\textbf{Anchor-free 2D object detection.}
The implementation of the back-tracing representative points in our \method adopts similar anchor-free localization strategies in 2D object detection.
Unlike two-stage 2D object detectors such as Faster RCNN~\cite{faster-rcnn}, SSD~\cite{ssd} and YOLOv2~\cite{yolov2} that generate proposals with the predefined anchors, the anchor-free detectors~\cite{cornernet,tychsen2017denet,zhou2019bottom,fcos,reppoints,redmon2016you,duan2019centernet}, especially the regression-based approaches~\cite{fcos,reppoints,kong2020foveabox,redmon2016you,duan2019centernet}, either directly regress borders~\cite{redmon2016you}, regress the object boundaries with an iterative dynamic sampling strategy~\cite{reppoints}, or regress 4D offsets as the surrogate of the localization results\cite{fcos}.
Inspired by these methods, in our method, the back-tracing process relies on a \emph{class-agnostic} offset regressor to retrieve the representative points that indicate the likely shape profile surrounding each vote center and thus provides more local structural clues for latter inference.
Rather than localization constrained by predefined \emph{class-aware} statistics, as in VoteNet~\cite{votenet} and its successors, the proposed \method benefits more flexible regression without losing its discriminative power.

\vspace{+1mm}
\noindent\textbf{Back-tracing in voting-based object detection and instance segmentation.}
Leibe~\etal~\cite{robust-hough} applied the hough voting strategy for simultaneously 2D object detection and instance segmentation.
The core part of this approach is a learned highly flexible representation for object shapes in a probabilistic extension of Generalized Hough Transform.
Moreover, the work in \cite{scene-cut} combined the top-down clues available from object detection and the bottom-up power of Markov Random Fields (MRFs) when performing class-specific object detection and segmentation in 3D scenes.
These methods rely on a top-down strategy such as back-tracing object hypotheses to enhance the bottom-up strategy such as Hough voting.
Their mutual agreement enhances each other, and thus devotes to the success of more reliable object detection.
The proposed \method also follows this idea with a new end-to-end trainable back-tracing process based on the representative points. 
Recently, as a 3D instance segmentation method, 3D-MPA~\cite{3d-mpa} applied a ``direct'' back-tracing strategy to cluster the surface points from the corresponding votes in one cluster.
In contrast, our method alleviates the inherent partial coverage and outlier issues from the ``generative'' back-tracing strategy.

\section{Methodology}

In this section, we describe the technical details of our \methodnospace. Sec.~\ref{sub:overview} presents an overview of our method.
In Sec.~\ref{sub:representative_point_generation} to Sec.~\ref{subsec:learning-objective}, we elaborate the network architecture and the learning objective of our \methodnospace.


\subsection{Overview}
\label{sub:overview}

As illustrated in Fig.~\ref{fig:method}, the input of our \method is a point cloud $\mP \in \mathbb{R}^{N\times3}$, with a 3D coordinate for each of the $N$ points.
Such an input typically comes from multi-view stereo (\eg ScanNet~\cite{scannet}) or depth sensors (\eg SUN RGB-D~\cite{sunrgb-d}).
The output is a collection of (oriented) bounding boxes $\cB$, each box $b \in \cB$ is associated with a predefined category label $l_b \in \cC$, a center $\vc_b = [c_b^x, c_b^y, c_b^z]^\top \in \mathbb{R}^3$ in a world coordinate system, the size of bounding box $\mathbf{s}_b = [s_b^x, s_b^y, s_b^z]^\top\in\mathbb{R}^3$, an orientation angle $\theta_b$ in the $xy$-plane of the same world coordinate system.

\method consists of four main modules: (1) vote generation and clustering, (2) back-traced representative points generation, (3) seed point revisiting, and (4) proposal refinement and classification followed by standard 3D NMS.
In the first module, we follow the same network and training strategy as in VoteNet~\cite{votenet} to generate the seed points, the votes and the vote clusters.
We will elaborate the other three modules in the following parts.

\subsection{Generating Back-traced Representative Points}
\label{sub:representative_point_generation}

The conventional back-tracing step of Hough voting for identifying object boundaries~\cite{robust-hough} is less reliable for amodal object detection from partial observations, as it just picks up seed points that contribute to the selected votes.
For example, in VoteNet~\cite{votenet}, these back-traced seed points can only capture local geometric area near the cluster center while containing the outliers from the cluttered background in the meantime.
VoteNet~\cite{votenet} circumvents this issue by removing the back-tracing step and using a PointNet-like set aggregation block just for votes, and then generates the object proposals and classifies them.
However, the aforementioned incompleteness issue and the outliers within the votes (delivered from the seed points) are clearly harmful for the detection task.
To this end, we argue that it is still beneficial to use back-tracing in point-based 3D object detection, but it requires a better tracing strategy to effectively find the representative seed points.
In contrast to the conventional back-tracing strategy, we propose a representative point generation (RPG) module to backwardly regress the \emph{virtually} generated representative points from the votes in a \emph{generative} manner.
The generated representative points are uniformly distributed within the potential 3D area of a candidate object, which can also indicate 3D object shapes when interacted with their actual surrounding seed points.

To be specific, the vote sampling and grouping block generates a set of vote cluster centers $\{ \vv_i \}_{i=1}^M$, where $\vv_i = [\vp_i^\top, \vf_i^\top]^\top$ with $\vp_i \in\mathbb{R}^3$ as the vote's geometric position in the 3D space and $\vf_i\in\mathbb{R}^C$ as its feature extracted from the preceding network, $M$ is the number of vote clusters.
Then, the RPG module generates a set of representative points for each vote cluster center.
Rather than directly sampling the 3D coordinates of these points, this module simultaneously predicts the tentative orientation $\theta_i \in [0, 2\pi]$ of the potential object, and regresses the offset distances $\vx_i\in\mathbb{R}^6$ from $\vv_i$ to the tentative object's surface in $6$ canonical directions (\ie front/back/left/right/up/down), and then uniformly samples distributed representative points $\cR_i = \{\vr_i^k = (x_i^k, y_i^k, z_i^k)\}_{k=1}^K$ along these directions (which are skewed by the predicted orientation) within the range of the offset distances.
$K$ is the number of representative points.
In this work, we sample $2$ uniformly distributed points within the range of each offset, thus $K=2\times6=12$ in total.

\vspace{+1mm}
\noindent\textbf{Network architecture and learning.}
The RPG module is implemented by using multi-layer perceptrons (MLP) with the ReLU activation function and batch normalization.
It takes the feature $\vf_i$ from the vote center $\vv_i$ as the input, and its output is the set $\{\vx_i, \theta_i \}$.
We employ $\exp(\cdot)$ to map any real number to (0, $\infty$) on the output of $\vx_i$.
This module is supervised by the ground-truth (GT) offsets as the vote center can be assigned to a GT object, \ie
\begin{equation}
L_\text{rep-off} = \frac{1}{M_\text{pos}} \sum_{i=1}^M \|\vx_i-\vx_i^*\|_\rho \cdot \mathbb{I}[\vv_i~\text{is positive}],
\end{equation}
where $\mathbb{I}[\vv_i~\text{is positive}]$ indicates whether the vote center $\vv_i$ is around a GT object center (within a radius of $0.3$).
$M_\text{pos}$ is the number of positive vote centers.
$\rho$ means smooth-$\ell_1$ norm.
And $\vx_i^*$ is the GT offsets from the vote center $\vv_i$ to the $6$ faces of the GT bounding box.
This module is also supervised by the GT orientation of the same GT object.
To better predict the orientation angle, we adopt the bin-based angle prediction scheme as in~\cite{f-pointnet}, which predicts a classification score for each orientation bin and a regression offset in each bin, and then uses the cross-entropy loss for orientation bins, and the smooth-$\ell_1$ loss for the regression offset.
We term the orientation loss as $L_\text{rep-ang}$.
Therefore, the final learning objective for this module is 
\begin{equation}
L_\text{rep} = \lambda L_\text{rep-off} + L_\text{rep-ang},
\end{equation}
where $\lambda=20$ is used to balance the two terms.

\subsection{Revisiting Seed Points}

By back-tracing the representative points $\cR_i$ in a generative manner from a vote center $\vv_i, i=1,\ldots,M$, we can roughly obtain the size and the position of a possible object in a class-agnostic way, but it still requires mutual consistency from the actual seed points in order to reliably generate the object proposals for more accurate object localization, bounding box estimation and object class prediction.
To be specific, we revisit the seed points $\{\mathbf{q}_j | \| \mathbf{q}_j - \mathbf{r}^k_i \| \leq \delta \}$ within a fixed radius ($\delta$=0.2 in the work) surrounding a back-traced representative point $\vr^k_i, k = 1,\ldots,K$, and aggregate the revisited seed features by using a PointNet-like block~\cite{pointnet}, denoted as $\tilde{\vg}_{\vr^k_i}$.
This process is similarly implemented as ROI-grid pooling proposed in PV-RCNN~\cite{pvrcnn}, but with a different griding and radius selection strategy.
%


Thereafter, to each vote center (or called proposal) $\vv_i$, the set of aggregated seed point features $\tilde{\vg}_{\vr^k_i}$ from each representative point $\vr^k_i$ can be further fused into a single feature $\tilde{\vg}_{\vv_i}$, which is implemented by concatenating $\{ \tilde{\vg}_{\vr_i^k} \}_{k=1}^K$ in a predefined order before being projected to a $128$-dimensional feature.
The predefined order should be consistent for each proposal, but different ordering strategies do not affect the performance.
The revisited seed features are summarized into $\tilde{\vg}_{\vv_i}$, which thus captures the local object-level features from the relatively precise raw point clouds instead of the predicted vote points.

\subsection{Proposal Refinement and Classification}

The back-traced representative point set $\cR_i$ helps to revisit the seed points and aggregate the local geometric clues from the potential object indicated by the vote center $\vv_i$.
The aggregated feature $\tilde{\vg}_{\vv_i}$ can be concatenated with the feature $\vf_i$ of the vote center $\vv_i$, and then refine the proposal and use for more discriminative object class prediction.
To this end, the fused feature $\tilde{\vf}_i = [\tilde{\vg}_{\vv_i}^\top, \vf_i^\top]^\top \in \mathbb{R}^{256}$ is fed into a shared MLP to predict the residuals $\Delta\vx_i$ and $\Delta\theta_i$ based on the preceding estimation results $\vx_i$ and $\theta_i$, and produce the final 
output set $\{\vx_i+\Delta\vx_i, \theta_i + \Delta\theta_i\}$.
Meanwhile, we predict the objectness score and the semantic classification score for each fused feature, similarly as in~\cite{votenet}.
Note that the final offsets $\vx_i + \Delta\vx_i$ can be reformulated as the bounding box size $\mathbf{s}_i = [s_i^x, s_i^y, s_i^z]^\top\in\mathbb{R}^3$ and the object center $\vc_i = [c_i^x, c_i^y, c_i^z]^\top\in\mathbb{R}^3$, by min-max clipping the final representative point set $\tilde{\cR}_i$ in the canonical coordinate.

\subsection{The Learning Objective}
\label{subsec:learning-objective}

In summary, the loss function of the entire framework of the newly proposed \method is defined as following:
\begin{multline}
L = L_\text{vote-reg} + \lambda_\text{obj-cls} L_\text{obj-cls} + \\ \lambda_\text{sem-cls} L_\text{sem-cls} + \lambda_{rep} L_\text{rep} + \lambda_\text{refine} L_\text{refine}
\end{multline}
Following the terms and label assignment strategy used in VoteNet~\cite{votenet}, the loss terms $L_\text{vote-reg}$, $L_\text{obj-cls}$, $L_\text{sem-cls}$ indicate the per-point vote regression loss, the objectness loss and the semantic classification loss, respectively. $L_\text{rep}$ is defined in Sec.~\ref{sub:representative_point_generation}.
$L_\text{refine}$ is used to supervise the residuals from the initial representative point sets to the final representative point sets:
\begin{multline}
    L_\text{refine} = \frac{1}{M_\text{pos}}\sum_{i=1}^M (\lambda\|\vx_i + \Delta \vx_i - \vx_i^{*}\|_\rho + \\ 
    \|(\theta_i + \Delta\theta_i - \theta_i^{*}\|_\rho) \cdot \mathbb{I}[\vv_i~\text{is positive}]
\end{multline}
$\rho$ denotes the smooth-$\ell_1$ norm. $\theta_i^{*}$ is the orientation angle of the ground-truth object bounding box. $L_\text{refine}$ is computed only on the positive vote clusters.
The weighting factors are $\lambda_\text{obj-cls}=1$, $\lambda_\text{sem-cls}=0.1$, $\lambda_\text{rep}=1$, $\lambda_\text{refine}=1$ and $\lambda=20$.

\section{Experiments}
\label{sec:experiments}

\subsection{Setups and Implementation Details}
\label{subsec:exp-setup}

\noindent\textbf{Datasets.} We evaluate our method on two large-scale indoor scene datasets, \ie SUN RGB-D~\cite{sunrgb-d} and ScanNet V2~\cite{scannet}.
SUN RGB-D consists of $10,355$ single-view indoor RGB-D images annotated with the oriented 3D bounding boxes and the semantic labels for $37$ categories.
The point clouds are converted from the depth maps based on the provided camera parameters.
The captured point clouds contain severe occlusions and holes, thus are challenging for 3D object detection.
%
ScanNet V2 is a 3D mesh dataset about $1,500$ 3D reconstructed indoor scenes.
It contains $18$ object categories with densely annotated axis-aligned bounding boxes.
The scans in the ScanNet V2 dataset are more complete with more objects than those in the SUN RGB-D dataset.
For both datasets, we use the same data preparation and training/validation split as in VoteNet~\cite{votenet}.

\vspace{+1mm}
\noindent\textbf{Input and data augmentation.} The input of our method is a point cloud randomly sub-sampled from the raw data of each dataset, \ie, $20,000$ points from a point cloud in the SUN RGB-D dataset, and $40,000$ points from a 3D mesh in the ScanNet V2 dataset.
We also include the height feature to each point.
To augment the training data, we add random flipping, rotating and scaling to the input point clouds, as the way employed by VoteNet~\cite{votenet}.

\begin{table*}[t]
\small

\caption{
3D object detection results on the ScanNet V2 validation set(left) and the SUN RGB-D V1 validation set(right).
Evaluation metric is average precision with 3D IOU thresholds as 0.25 and 0.50.
*Note for fair comparison, we report the results of H3DNet on the ScanNet V2 dataset under both $1$ and $4$ PointNet++ backbones (BB) settings.
While we only report the result of H3DNet with $4$ PointNet++ backbones (BB) on the SUN RGB-D dataset, as the work~\cite{h3dnet} only reports the result under this setting.
}
\label{tab:sota-table}
\vspace{+1mm}
\begin{minipage}[t]{0.5\linewidth}
\centering
\resizebox{\textwidth}{22mm}{
\begin{tabular}{l|c|c|c}
\hline
 ScanNet V2 & Input & mAP@0.25 & mAP@0.50 \\ \hline
DSS~\cite{dss}  & Geo + RGB  & 15.2 & 6.8 \\
F-PointNet~\cite{f-pointnet}  & Geo + RGB  & 19.8 & 10.8 \\
GSPN~\cite{gspn}  & Geo + RGB  & 30.6 & 17.7 \\
3D-SIS~\cite{3d-sis}  & Geo + 5 views  & 40.2 & 22.5 \\
\hline
VoteNet~\cite{votenet} & Geo only & 58.6 & 33.5 \\
HGNet~\cite{hgnet} & Geo only & 61.3 & 34.4 \\
MLCVNet~\cite{mlcvnet} & Geo only & 64.7 & 42.1 \\
H3DNet (1BB)*~\cite{h3dnet} & Geo only & 64.4 & 43.4 \\
H3DNet (4BB)*~\cite{h3dnet} & Geo only & \textbf{67.2} & 48.1 \\
\hline
Ours & Geo only & 66.1 & \textbf{50.9} \\
\hline
\end{tabular}}
\end{minipage}
\begin{minipage}[t]{0.5\linewidth}
\centering

\resizebox{\textwidth}{22mm}{
\begin{tabular}{l|c|c|c}
\hline
SUN RGB-D & Input & mAP@0.25 & mAP@0.50 \\ \hline
DSS~\cite{dss} & Geo + RGB & 42.1 & - \\
COG~\cite{cog} & Geo + RGB & 47.6 & - \\
2D-driven~\cite{2d-driven} & Geo + RGB & 45.1 & - \\
F-PointNet~\cite{f-pointnet} & Geo + RGB & 54.0 & - \\
\hline
VoteNet~\cite{votenet} & Geo only & 57.7 & 32.9 \\
HGNet~\cite{hgnet} & Geo only & \textbf{61.6} & - \\
MLCVNet~\cite{mlcvnet} & Geo only & 59.8 & - \\
H3DNet (1BB)*~\cite{h3dnet} & Geo only & - & - \\
H3DNet (4BB)*~\cite{h3dnet} & Geo only & 60.1 & 39.0 \\
\hline
Ours & Geo only & 61.1 & \textbf{43.7} \\
\hline
\end{tabular}}
\end{minipage}
\end{table*}

\begin{table*}[t]
\caption{
3D object detection results on the ScanNet V2 validation set. 
The evaluation metric is the average precision with 3D IOU threshold as 0.50.
*Note that for H3DNet only the per-category results with $4$ PointNet++ backbones are reported in \cite{h3dnet}.
}
\label{tab:scannet-per-class}
\vspace{+1mm}
\resizebox{\textwidth}{!}{
\centering
\begin{tabular}{c|c c c c c c c c c c c c c c c c c c|c}
\hline
ScanNet V2 & cab & bed & chair & sofa & tabl & door & wind & bkshf & pic & cntr & desk & curt & fridg & showr & toil & sink & bath & ofurn & avg \\
\hline
VoteNet~\cite{votenet} & 8.1 & 76.1 & 67.2 & 68.8 & 42.4 & 15.3 & 6.4 & 28.0 & 1.3 & 9.5 & 37.5 & 11.6 & 27.8 & 10.0 & 86.5 & 16.8 & 78.9 & 11.7 & 33.5 \\
MLCVNet~\cite{mlcvnet} & 16.6 & \textbf{83.3} & 78.1 & 74.7 & 55.1 & 28.1 & 17.0 & \textbf{51.7} & 3.7 & 13.9 & 47.7 & 28.6 & 36.3 & 13.4 & 70.9 & 25.6 & 85.7 & 27.5 & 42.1 \\
H3DNet*~\cite{h3dnet} & 20.5 & 79.7 & 80.1 & 79.6 & 56.2 & 29.0 & 21.3 & 45.5 & 4.2 & 33.5 & 50.6 & 37.3 & 41.4 & \textbf{37.0} & \textbf{89.1} & 35.1 & \textbf{90.2} & \textbf{35.4} & 48.1 \\
\hline
Ours & \textbf{28.7} & 80.6 & \textbf{81.9} & \textbf{80.6} & \textbf{60.8} & \textbf{35.5} & \textbf{22.2} & 48.0 & \textbf{7.5} & \textbf{43.7} & \textbf{54.8} & \textbf{39.1} & \textbf{51.8} & 35.9 & 88.9 & \textbf{38.7} & 84.4 & 33.0 & \textbf{50.9} \\
\hline
\end{tabular}}
\end{table*}

\vspace{+1mm}
\noindent\textbf{Network training details.}
Our network is end-to-end optimized by using the Adam optimizer with the batch size as $8$.
The base learning rates are $0.001$ for the SUN RGB-D~\cite{sunrgb-d} dataset and $0.005$ for the ScanNet V2~\cite{scannet} dataset. We train the network for $220$ epochs on both datasets. 
The cosine annealing learning rate strategy\cite{loshchilov2016sgdr} is adopted for learning rate decay.
Based on PyTorch platform equipped with one NVIDIA GeForce RTX 2080 Ti GPU card, it takes around $4$ hours to train the model on the ScanNet V2 dataset, while it takes around 12 hours on the SUN RGB-D dataset.

\vspace{+1mm}
\noindent\textbf{Inference and evaluation.}
Our method takes the point clouds of the entire scenes as the inputs and outputs the object proposals. 
The proposals are post-processed by a 3D NMS module with an IoU threshold of $0.25$.
The evaluation follows the same protocol as in~\cite{dss} using mean average precision, especially mAP@$0.25$ and mAP@$0.50$.

\subsection{Comparisons with the State-of-the-art Methods}
\label{subsec:exp-results}

We compare our method with a list of reference methods, for example the earlier attempts, such as COG~\cite{cog}, DSS~\cite{dss} and 3D-SIS~\cite{3d-sis}, 2D-driven~\cite{2d-driven} and F-PointNet~\cite{f-pointnet}, and GSPN~\cite{gspn}, and the recent point cloud-based state-of-the-art methods such as VoteNet~\cite{votenet} and its successors MLCVNet~\cite{mlcvnet}, HGNet~\cite{hgnet} and H3DNet~\cite{h3dnet}.

\begin{figure*}
\centering
\includegraphics[width=0.9\linewidth]{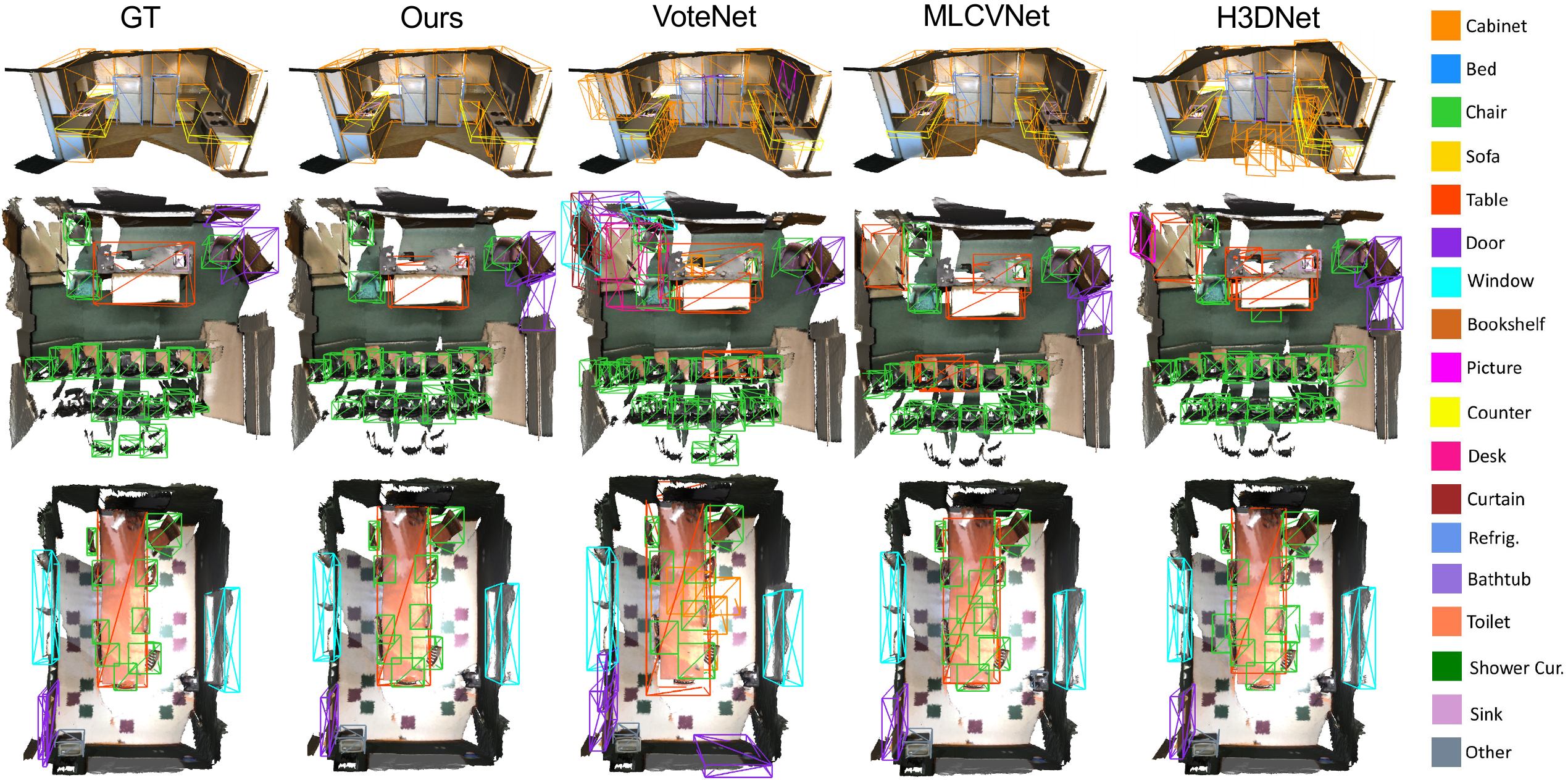}
\caption{
Qualitative results of different 3D object detection methods on ScanNet V2 dataset~\cite{scannet}. 
The baseline methods are VoteNet~\cite{votenet}, MLCVNet~\cite{mlcvnet} and H3DNet~\cite{h3dnet}. 
Best viewed on screen.
}
\vspace{-5mm}
\label{fig:vis_results_scannet}
\end{figure*}

\vspace{+1mm}
\noindent\textbf{Quantitative results.}
The comparison results are summarized in Table~\ref{tab:sota-table}.
Our method outperforms all baseline methods by remarkable performance gains, for example more than $7.5$\% and $4.7$\% improvement in terms of the mAP@$0.50$ metric on the validation sets of ScanNet V2 and SUN RGB-D respectively.
Note that mAP@$0.50$ is a fairly challenging metric as it basically requires more than $79\%$ coverage in each dimension of a bounding box, which indicates that back-tracing representative points can  significantly improve the localization accuracy.
Notably, MLCVNet~\cite{mlcvnet} works well on the ScanNet dataset but achieves relatively poor performance on the SUN RGB-D dataset, while HGNet~\cite{hgnet} works well on the SUN RGB-D dataset but achieves poor result on the ScanNet dataset, especially in terms of the mAP@$0.50$ metric.
Our method works well on both datasets, which indicates its stronger generalization ability for different detection scenarios. 
ScanNet contains relative complete 3D reconstructed meshes, while SUN RGB-D consists of single-view RGB-D scans with severe occlusions and holes.
Moreover, H3DNet~\cite{h3dnet} ensembles $4$ PointNet++~\cite{qi2017pointnet++} backbones to achieve the reported result on the SUN RGB-D dataset, while our model only needs one backbone as the base feature extractor.
It further validates it is effective to back-trace the representative points for reliably parsing the object proposals.
As shown in Table~\ref{tab:scannet-per-class}, our method performs the best on $12$ classes among $18$ total classes from the ScanNet dataset in terms of mAP@$0.50$.
While our method only uses one PointNet++ backbone for point cloud feature extraction, it outperforms H3DNet~\cite{h3dnet} with $4$ PointNet++ backbones.
Moreover, it achieves better performance on the categories (\eg ``cabinet'', ``chair'', ``sofa'', ``table'', ``counter'' and ``desk'') with irregular sizes or shapes, as its back-tracing and revisiting process removes the outliers from the votes and enables better mutual agreement between the votes and the local object surfaces, whilst its class-agnostic regression strategy makes the estimation process robust to shape variations.

\begin{figure*}
\centering
\includegraphics[width=0.9\linewidth]{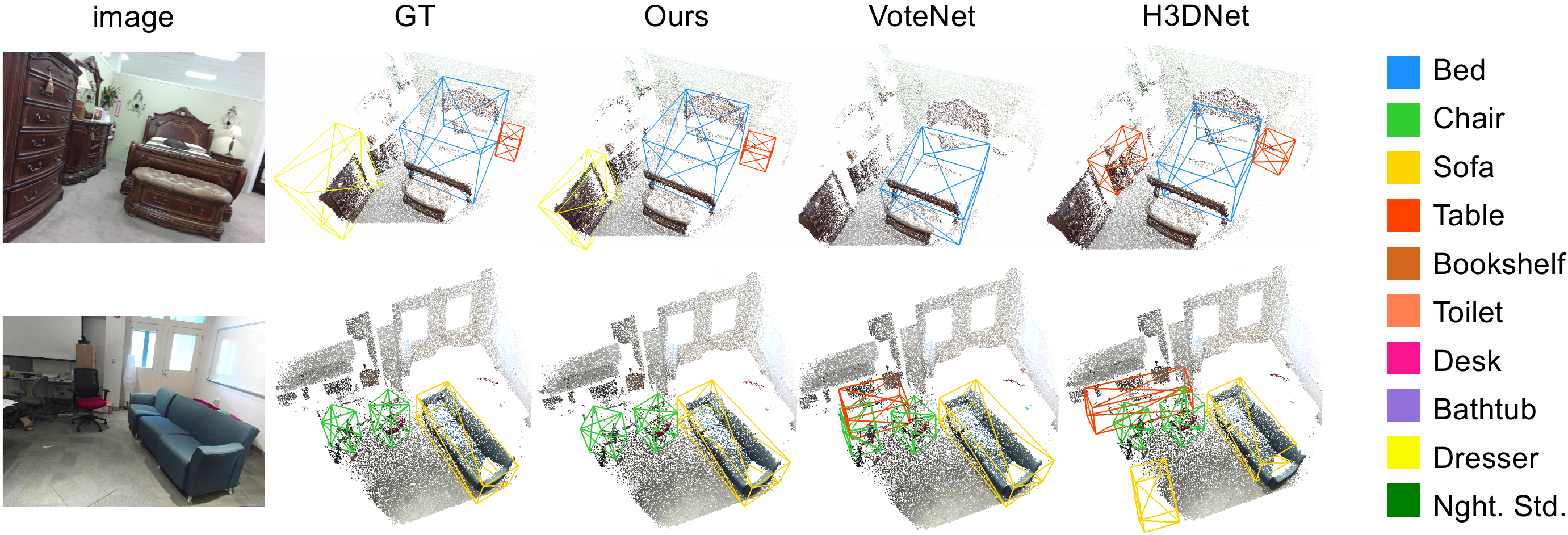}
\caption{
Qualitative comparison results of the 3D object detection methods on the SUN RGB-D dataset~\cite{sunrgb-d}. 
The baseline methods are VoteNet~\cite{votenet} and H3DNet~\cite{h3dnet}. 
Best viewed on screen.
}
\label{fig:vis_results_sunrgbd}
\vspace{-5mm}
\end{figure*}

\vspace{+1mm}
\noindent\textbf{Qualitative results.}
In Fig.~\ref{fig:vis_results_scannet} and Fig.~\ref{fig:vis_results_sunrgbd}, we visualize the representative 3D object detection results, from our method and the baseline methods, such as VoteNet~\cite{votenet}, MLCVNet~\cite{mlcvnet} and H3DNet~\cite{h3dnet}.
These results demonstrate that our method achieves more reliable detection results with more accurate bounding boxes and orientations. Our method also eliminates false positives and discovers more missing objects when compared with the baseline methods\footnote{MLCVNet does not provide a checkpoint for the SUN RGB-D dataset~\cite{sunrgb-d} thus we cannot provide its visualization results on this dataset.}.

\begin{figure}
    \centering
    \includegraphics[width=0.8\linewidth]{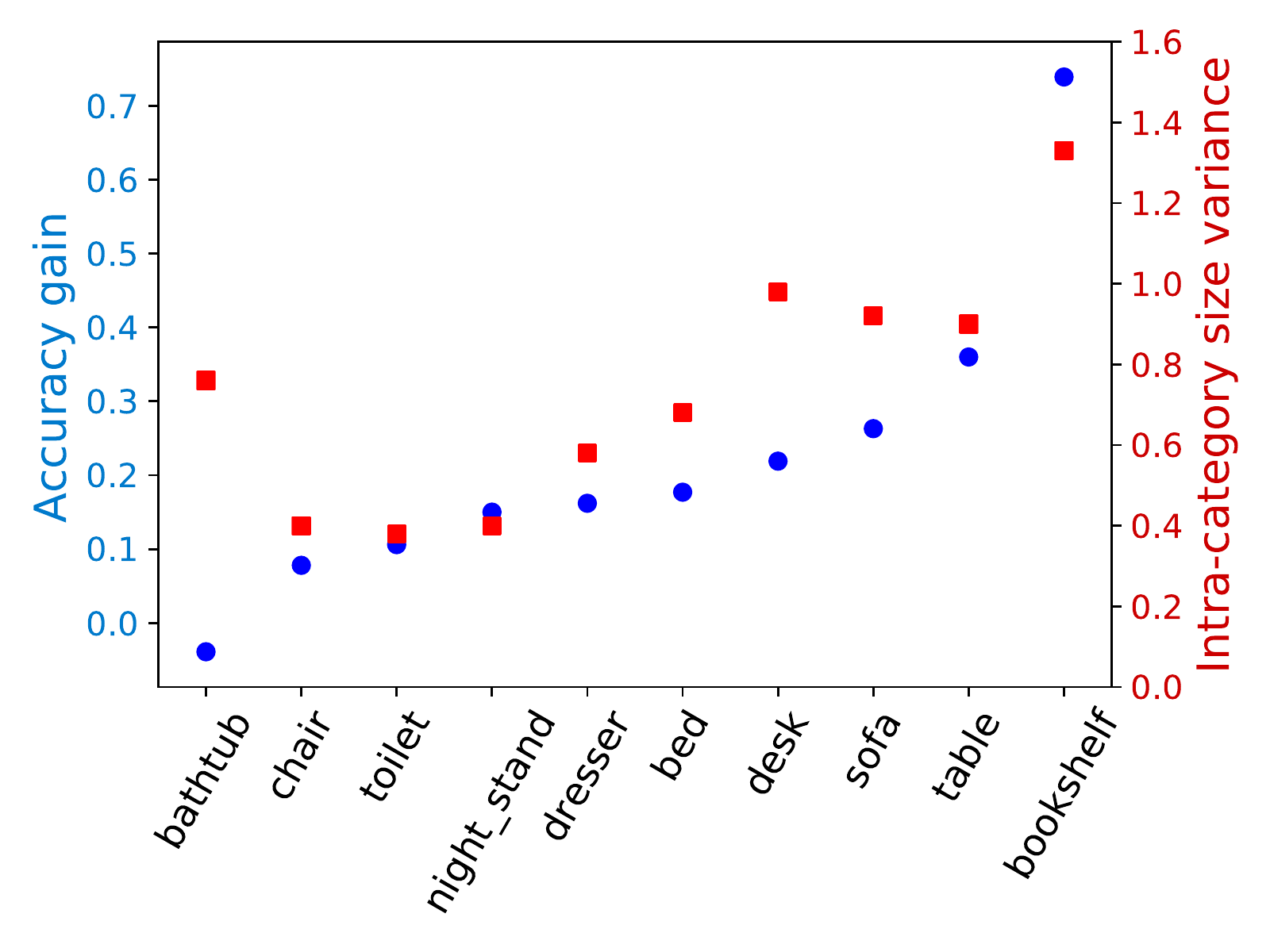}
    \vspace{-2mm}
    \caption{
Class-agnostic bounding box regression works better on the categories with high intra-category size variances. 
For each category we show the relative accuracy gain (in blue dots) of the alternative method ``VoteNet+CA-Reg'' over VoteNet~\cite{votenet} and intra-category size variance(in red squares), which is normalized by the mean category size.
}
    \vspace{-5mm}
    \label{fig:AP_gain}
\end{figure}

\subsection{Ablation Study and Discussions}
\label{subsec:exp-analysis}

\noindent\textbf{Class-agnostic bounding box regression.}
Our method regresses the representative points in a class-agnostic way, which are then converted to the proposal's bounding boxes.
Note VoteNet~\cite{votenet} and its variants~\cite{hgnet,mlcvnet,h3dnet} have to estimate the sizes of object proposals in a class-aware way.
Thus these baseline methods usually output the object sizes that can only moderately vary around the class-aware templates, and tend to falsely detect the objects when their sizes are unusual.
To validate this observation, we implement an alternative method that employs a similar regression strategy as in our method but shares the same network as VoteNet~\cite{votenet}.
We term this variant as ``VoteNet+CA-Reg''.
As shown in Table~\ref{tab:ablation}, this variant significantly outperforms VoteNet.
As shown in Figure~\ref{fig:AP_gain}, we also observe that this alternative method works better for the categories with high intra-category variance in sizes, and the mAP@$0.50$ gains of this alternative method over VoteNet on the SUN RGB-D dataset are positively related to size variances.

\begin{table}[t]
\centering
\caption{Quantitative ablation experiments on ScanNet V2 and SUN RGB-D datasets. ``+CA-Reg'' means VoteNet~\cite{votenet} with a class-agnostic bounding box regressor, ``+Seed-Pts'' indicates VoteNet with votes fused with their corresponding seed points.}
\label{tab:ablation}
\vspace{+1mm}
\resizebox{\linewidth}{!}{
        \begin{tabular}{l|c|c|c|c}
        \hline
 & \multicolumn{2}{c|}{ScanNet V2} & \multicolumn{2}{c}{SUN RGB-D} \\
 \cline{2-5}
 & mAP@$0.25$ & mAP@$0.50$ & mAP@$0.25$ & mAP@$0.50$ \\
 \hline
 VoteNet & 58.6 & 33.5 & 57.7 & 32.9 \\
  +CA-Reg & 59.3 & 40.8 & 58.2 & 37.6 \\
 +Seed-Pts & 59.1 & 37.6 & 59.5 & 33.6\\
 \hline
Ours & \textbf{66.1} & \textbf{50.9} & \textbf{61.1} & \textbf{43.7} \\
 \hline
\end{tabular}}
\vspace{-2mm}
\end{table}

\begin{figure}[t]
    \centering
    \subfigure[Corresponding seed points]{
        \includegraphics[width=0.47\linewidth]{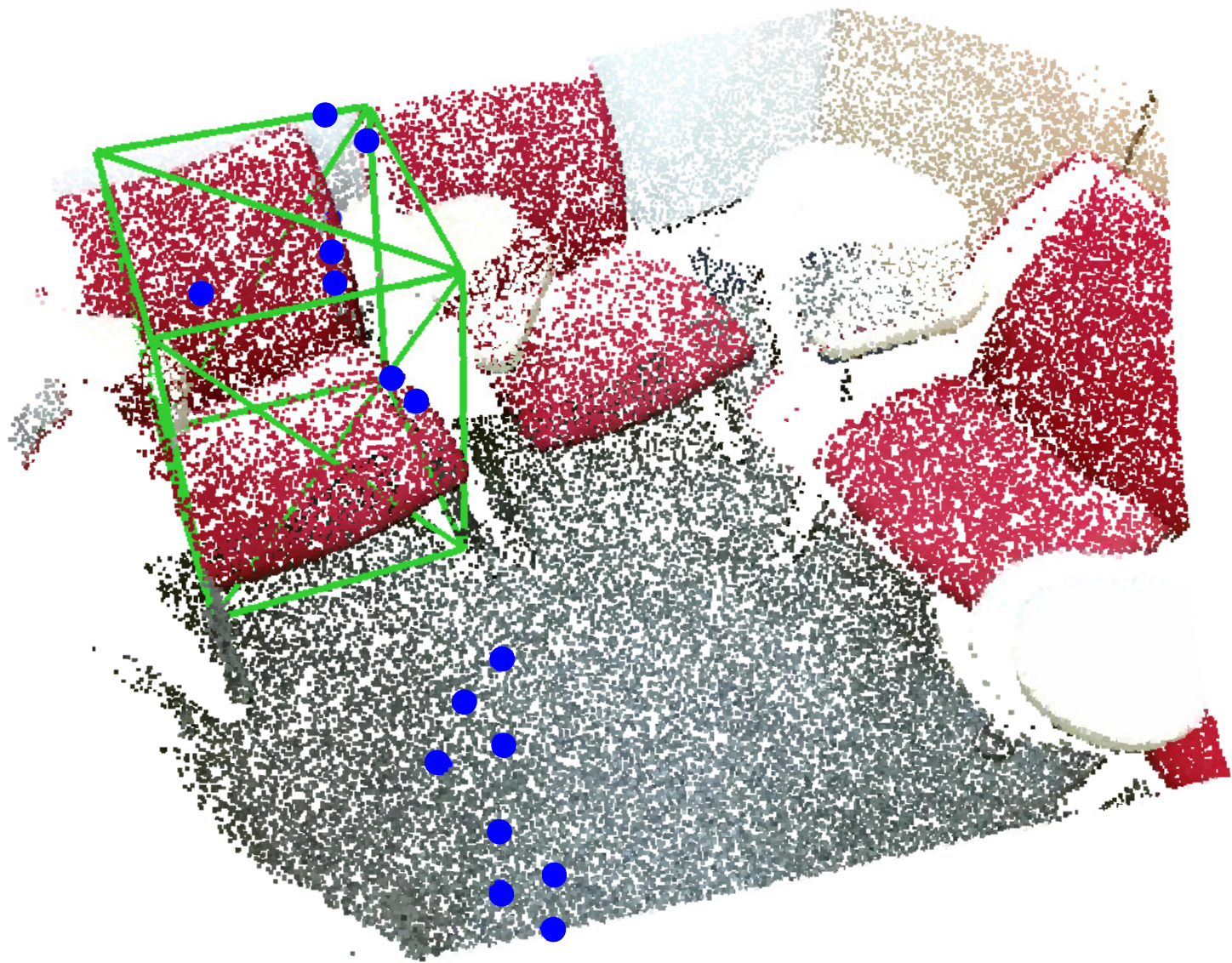}
    }
    \subfigure[Revisited seed points]{
        \includegraphics[width=0.47\linewidth]{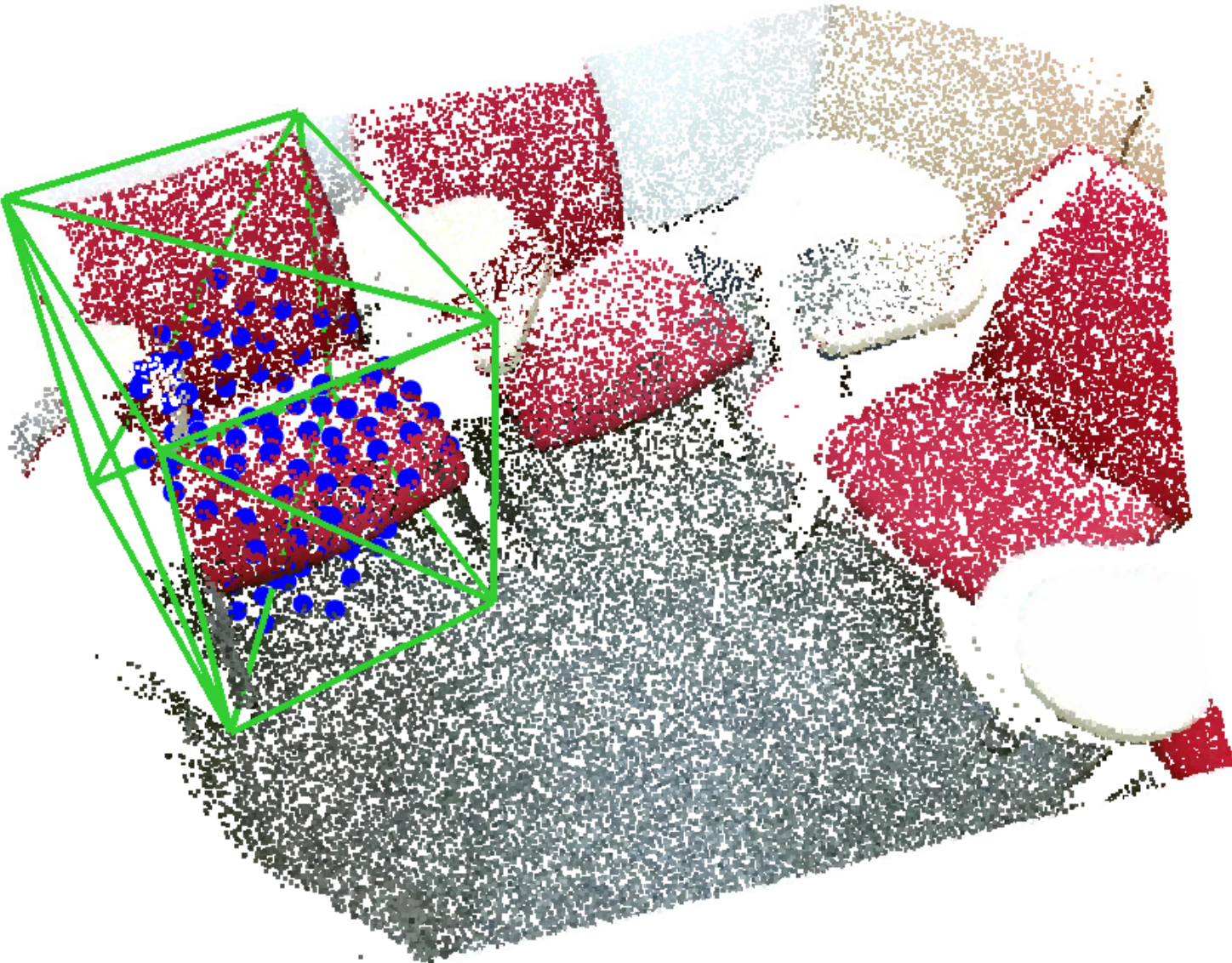}
    }
    \caption{
    Comparison between the corresponding seed points and the revisited seed points. 
    The seed points are marked as blue points and the predicted bounding boxes are the green boxes. 
    The revisited seed points completely cover the chair, while the corresponding seed points suffer from partial coverage and the outliers.
    }
    \vspace{-3mm}
    \label{fig:revisited-seed-points}
\end{figure}

\vspace{+1mm}
\noindent\textbf{Back-tracing, revisiting and refinement.}
Back-tracing the representative points should also be combined with the subsequent revisiting and refinement modules.
As shown in Table~\ref{tab:ablation}, we find this complete method has significant performance gains ($\sim10$\% mAP improvement on ScanNet and $\sim6$\% mAP improvement on SUN RGB-D in terms of mAP@$0.50$) over the aforementioned baseline.
The back-tracing operation gives rough estimation of the object extent, and the revisiting and refining operations further update the proposal features with the reliable seed features in the neighborhood, thus offering better chance to produce more accurate detection results.
Moreover, as shown in Figure~\ref{fig:revisited-seed-points}, the revisited seed points by our method compactly cover the object's surface, while the corresponding seed points retrieved by the votes can only partially cover the surface, and also suffer from the outliers. 

Moreover, to validate whether the seed points can help improve the object detection results, we consider another variant (termed as ``VoteNet+Seed-Pts'') that VoteNet has its vote features fused with the corresponding seed points' features.
In comparison to VoteNet, this alternative method also achieves non-trivial gains on both datasets, especially on ScanNet V2 in terms of mAP@$0.50$.

\vspace{+1mm}
\noindent\textbf{Sampling strategy of representative points.}
In Table~\ref{tab:rep_conf}, we compare different sampling strategies to generate our representative points.
``Ray'' means uniform sampling along $6$ directions between $0$ and the maximum offsets. 
``Grid'' means uniform sampling within the 3D bounding box spanned based on the predicted offsets.
``\#Pts'' is the number of sampled points.
Our methods using different strategies are generally comparable. 

\vspace{+1mm}
\noindent\textbf{Model size and speed.}
As listed in Table~\ref{tab:model-size}, our proposed method is efficient in comparison to VoteNet, and is $3\times$ faster than the current state-of-the-art H3DNet~\cite{h3dnet}, when evaluated on both datasets.
Its model size is marginally increased from that of VoteNet, and around $4\times$ smaller than that of H3DNet.
Knowing that the proposed method has significant performance gains than these reference methods (as discussed in Sec.~\ref{subsec:exp-results}), its lightweight model validates that the proposed back-tracing strategy is significant for 3D object detection in point clouds\footnote{Note that MLCVNet does not provide a checkpoint for the SUN RGB-D dataset, we omit its comparison on this dataset.}.

\vspace{+1mm}
\noindent\textbf{Number of Backbones.}
Our \method can also be improved after using $4$ backbones, and it achieves the result of 51.8\% in terms of mAP@$0.50$ on ScanNet~\cite{scannet}, which outperforms H3DNet ($4$ backbones) with a remarkable margin (+3.7\%).

\begin{table}[t]
   \vspace{-2mm}
   \footnotesize
   \centering
   \caption{Results of \method using different RP sampling strategies.}
       \resizebox{\linewidth}{!}{
           \begin{tabular}{c|c|c|c|c|c}
           \hline
   & & \multicolumn{2}{c|}{ScanNet V2} & \multicolumn{2}{c}{SUN RGB-D} \\
   \cline{3-6}
     Types & \#Pts & mAP@$0.25$ & mAP@$0.50$ & mAP@$0.25$ & mAP@$0.50$ \\
    \hline
    Ray  &  6 & 65.0 & 48.3 & 60.3 & 42.7 \\
    Ray  & 12 & \textbf{66.1}  & \textbf{50.9} & \textbf{61.1} & \textbf{43.7} \\
    Ray  & 18 & 65.8 & 48.4 & 60.4 & 42.9 \\
    \hline
    Grid &  8 & 65.4 & 49.1 & 59.9 & 42.2 \\
    Grid & 27 & 66.0 & 49.2 & 60.2 & 42.5 \\
    \hline
    \end{tabular}}
    \label{tab:rep_conf}
    \vspace{-2mm}
   \end{table}

\begin{table}[t]
\caption{
Model size and processing time comparison of different methods, 
which are evaluated on a NVIDIA GeForce RTX 2080 Ti GPU card with the same configuration.
\#BB means the number of backbones used for feature extraction.
}
\label{tab:model-size}
\vspace{+1mm}
\centering
\resizebox{\linewidth}{!}{
\begin{tabular}{c|c|c|c|c}
     \hline
     Method & \#BB & Model size & ScanNet & SUN RGB-D \\
     \hline
     VoteNet~\cite{votenet} & 1 & 11.2MB & 0.130s & 0.076s \\
     MLCVNet~\cite{mlcvnet} & 1 & 13.9MB & 0.141s & - \\
     H3DNet~\cite{h3dnet} & 4 & 56.0MB & 0.330s & 0.241s \\
     \hline
     Ours & 1 & 12.9MB & 0.133s & 0.079s \\
     \hline
\end{tabular}}
\vspace{-5mm}
\end{table}

\section{Conclusion}

In this work, we have introduced a new approach to improve the voting-based 3D object detection method by generatively and class-agnostically back-tracing the representative points.
We revisit the seed points around the back-traced representative points and extract fine object surface features to generate the high-quality object proposals.
Comprehensive ablation studies show the importance and effectiveness of the proposed back-tracing, revisiting and refinement operations. 
Qualitative and quantitative results further demonstrate that our method remarkably outperforms the existing methods while bringing negligible increases in model size and executive time compared with VoteNet~\cite{votenet}.

\vspace{+1mm}
\noindent\textbf{Acknowledgements.} This work was supported by Key Research and Development Program of Guangdong Province, China, under Grant No. 2019B010154003, and the National Natural Science Foundation of China under Grant No. 61906012. We thank Zizheng Que and Zinuo You for valuable discussions and feedback.

{\small
\bibliographystyle{ieee_fullname}
\bibliography{egbib}
}

\clearpage


\setcounter{table}{0}
\setcounter{figure}{0}
\setcounter{section}{0}
\renewcommand\thesection{\Alph{section}}
\renewcommand{\thefigure}{S\arabic{figure}}
\renewcommand{\thetable}{S\arabic{table}}

\begin{table}[b]
    \caption{3D object detection results on ScanNetV2 dataset with multiple IoU thresholds. *Note that H3DNet~\cite{h3dnet} only provide the checkpoint with $4$ PointNet++ backbones as we use here.}
    \vspace{+2mm}
    \resizebox{\linewidth}{!}{
    \begin{tabular}{c|c|c|c}
        \hline
        ScanNet V2 & mAP@0.25 & mAP@0.50 & mAP@0.75 \\
        \hline
        VoteNet~\cite{votenet} & 58.6 & 33.5 & 3.4 \\
        HGNet~\cite{hgnet} & 61.3 & 34.4 & - \\
        MLCVNet~\cite{mlcvnet} & 64.7 & 42.1 & 7.4 \\
        H3DNet*~\cite{h3dnet} & \textbf{67.2} & 48.1 & 15.4 \\
        \hline
        Ours & 66.1 & \textbf{50.9} & \textbf{19.1} \\
        \hline
    \end{tabular}}
    \label{scannet@0.75}
\end{table}

\begin{table}[b]
    \caption{3D object detection results on SUN RGB-D dataset with multiple IoU thresholds. *Note that H3DNet~\cite{h3dnet} only provide the checkpoint with $4$ PointNet++ backbones as we use here. Also H3DNet use $40,000$ points for SUN RGB-D dataset as input, while others use $20,000$ points. }
    \vspace{+2mm}
    \resizebox{\linewidth}{!}{
    \begin{tabular}{c|c|c|c}
        \hline
        SUN RGB-D & mAP@0.25 & mAP@0.50 & mAP@0.75 \\
        \hline
        VoteNet~\cite{votenet} & 57.7 & 32.9 & 1.2 \\
        HGNet~\cite{hgnet} & \textbf{61.6} & - & - \\
        MLCVNet~\cite{mlcvnet} & 59.8 & - & - \\
        H3DNet*~\cite{h3dnet} & 60.1 & 39.0 & 3.5 \\
        \hline
        Ours & 61.1 & \textbf{43.7} & \textbf{5.3}\\
        \hline
    \end{tabular}}
    \label{sunrgbd@0.75}
\end{table}

\begin{table*}[t]
\caption{3D object detection results on ScanNet V2 validation dataset. Evaluation metric is the average precision with 3D IOU threshold $0.25$. *Note that H3DNet~\cite{h3dnet} only reports the per-category results with $4$ PointNet++ backbones, while others with only $1$ PointNet++ backbone. }
    \resizebox{\textwidth}{!}{ 
    \begin{tabular}{c|c c c c c c c c c c c c c c c c c c|c}
    \hline
    ScanNet V2 & cab & bed & chair & sofa & tabl & door & wind & bkshf & pic & cntr & desk & curt & fridg & showr & toil & sink & bath & ofurn & mAP@0.25 \\
    \hline
    VoteNet\cite{votenet} & 36.3 & 87.9 & 88.7 & 89.6 & 58.8 & 47.3 & 38.1 & 44.6 & 7.8 & 56.1 & 71.7 & 47.2 & 45.4 & 57.1 & 94.9 & 54.7 & 92.1 & 37.2 & 58.7 \\
    MLCVNet\cite{mlcvnet} & 44.6 & \textbf{89.6} & 91.4 & 87.2 & 67.1 & 56.8 & 45.9 & \textbf{59.5} & 15.1 & 56.7 & 74.3 & 53.4 & 54.7 & 73.1 & 97.8 & 55.6 & 91.3 & 50.9 & 64.7 \\
    H3DNet*\cite{h3dnet} & \textbf{49.4} & 88.6 & 91.8 & \textbf{90.2} & 64.9 & \textbf{61.0} & \textbf{51.9} & 54.9 & \textbf{18.6} & 62.0 & 75.9 & \textbf{57.3} & 57.2 & \textbf{75.3} & \textbf{97.9} & \textbf{67.4} & \textbf{92.5} & \textbf{53.6} & \textbf{67.2} \\
    \hline
    Ours & 49.3 & 88.3 & \textbf{91.9} & 86.9 & \textbf{69.3} & 59.2 & 45.9 & 52.1 & 15.3 & \textbf{72.0} & \textbf{76.8} & 57.1 & \textbf{60.4} & 73.6 & 93.8 & 58.8 & 92.2 & 47.1 & 66.1 \\
    \hline
    \end{tabular}}
    \label{scannet@0.25}
\end{table*}

\begin{table*}[t]
\caption{3D object detection results on SUN RGB-D val dataset. We show per-category results of average precision (AP) with 3D IOU threshold $0.25$ as proposed in ~\cite{sunrgb-d}, and mean of AP across all semantic classes. For fair comparison with previous methods, the evaluation is on the SUN RGB-D V1 data. *Note that H3DNet~\cite{h3dnet} sub-samples $40,000$ points from every scene in SUN RGB-D dataset, while others use $20,000$ points. Also, H3DNet~\cite{h3dnet} only reports the per-category results with $4$ PointNet++ backbones, while others with only $1$ PointNet++ backbone.}
\vspace{+2mm}
\resizebox{\textwidth}{!}{
\begin{tabular}{c|c c c c c c c c c c|c}
    \hline
    SUN RGB-D & bathtub & bed & bookshelf & chair & desk & dresser & nightstand & sofa & table & toilet & mAP@0.25 \\
    \hline
    DSS\cite{dss} & 44.2 & 78.8 & 11.9 & 61.2 & 20.5 & 6.4 & 15.4 & 53.5 & 50.3 & 78.9 & 42.1 \\
    COG\cite{cog} & 58.3 & 63.7 & 31.8 & 62.2 & \textbf{45.2} & 15.5 & 27.4 & 51.0 & 51.3 & 70.1 & 47.6 \\
    2D-driven\cite{2d-driven} & 43.5 & 64.5 & 31.4 & 48.3 & 27.9 & 25.9 & 41.9 & 50.4 & 37.0 & 80.4 & 45.1 \\
    F-PointNet\cite{f-pointnet} & 43.3 & 81.1 & 33.3 & 64.2 & 24.7 & 32.0 & 58.1 & 61.1 & 51.1 & 90.9 & 54.0 \\
    \hline
    VoteNet\cite{votenet} & 74.4 & 83.0 & 28.8 & 75.3 & 22.0 & 29.8 & 62.2 & 64.0 & 47.3 & 90.1 & 57.7 \\
    MLCVNet\cite{mlcvnet} & \textbf{79.2} & 85.8 & 31.9 & 75.8 & 26.5 & 31.3 & 61.5 & 66.3 & 50.4 & 89.1 & 59.8 \\
    H3DNet*\cite{h3dnet} & 73.8 & 85.6 & 31.0 & 76.7 & 29.6 & 33.4 & 65.5 & \textbf{66.5} & 50.8 & 88.2 & 60.1 \\
    HGNet\cite{hgnet} & 78.0 & 84.5 & \textbf{35.7} & 75.2 & 34.3 & \textbf{37.6} & 61.7 & 65.7 & 51.6 & 91.1 & \textbf{61.6} \\
    \hline
    Ours & 76.2 & \textbf{86.9} & 29.7 & \textbf{77.4} & 29.6 & 35.9 & \textbf{65.9} & 66.4 & \textbf{51.8} & \textbf{91.3} & 61.1 \\
    \hline
\end{tabular}}
\label{sunrgbd@0.25}
\end{table*}

\begin{table*}[t]
\caption{3D object detection results on SUN RGB-D val dataset. We show per-category results of average precision (AP) with 3D IOU threshold $0.50$ as proposed in ~\cite{sunrgb-d}, and mean of AP across all semantic classes. For fair comparison with previous methods, the evaluation is on the SUN RGB-D V1 data. *Note that H3DNet~\cite{h3dnet} sub-samples $40,000$ points from every scene in SUN RGB-D dataset, while others use $20,000$ points. Also, H3DNet~\cite{h3dnet} only reports the per-category results with $4$ PointNet++ backbones, while others with only $1$ PointNet++ backbone.}
\vspace{+2mm}
\resizebox{\textwidth}{!}{
    \begin{tabular}{c|c c c c c c c c c c|c}
    \hline
    SUN RGB-D & bathtub & bed & bookshelf & chair & desk & dresser & nightstand & sofa & table & toilet & mAP@0.50 \\
    \hline
    VoteNet\cite{votenet} & 49.9 & 47.3 & 4.6 & 54.1 & 5.2 & 13.6 & 35.0 & 41.4 & 19.7 & 58.6 & 32.9 \\
    H3DNet*\cite{h3dnet} & 47.6 & 52.9 & 8.6 & 60.1 & 8.4 & 20.6 & 45.6 & 50.4 & 27.1 & 69.1 & 39.0 \\
    \hline
    Ours & \textbf{55.5} & \textbf{63.8} & \textbf{9.3} & \textbf{61.6} & \textbf{10.0} & \textbf{27.3} & \textbf{53.2} & \textbf{56.7} & \textbf{28.6} & \textbf{70.9} & \textbf{43.7} \\
    \hline
\end{tabular}}    
\label{sunrgbd@0.50}
\end{table*}

\begin{figure}[p]
    \includegraphics[width=\linewidth]{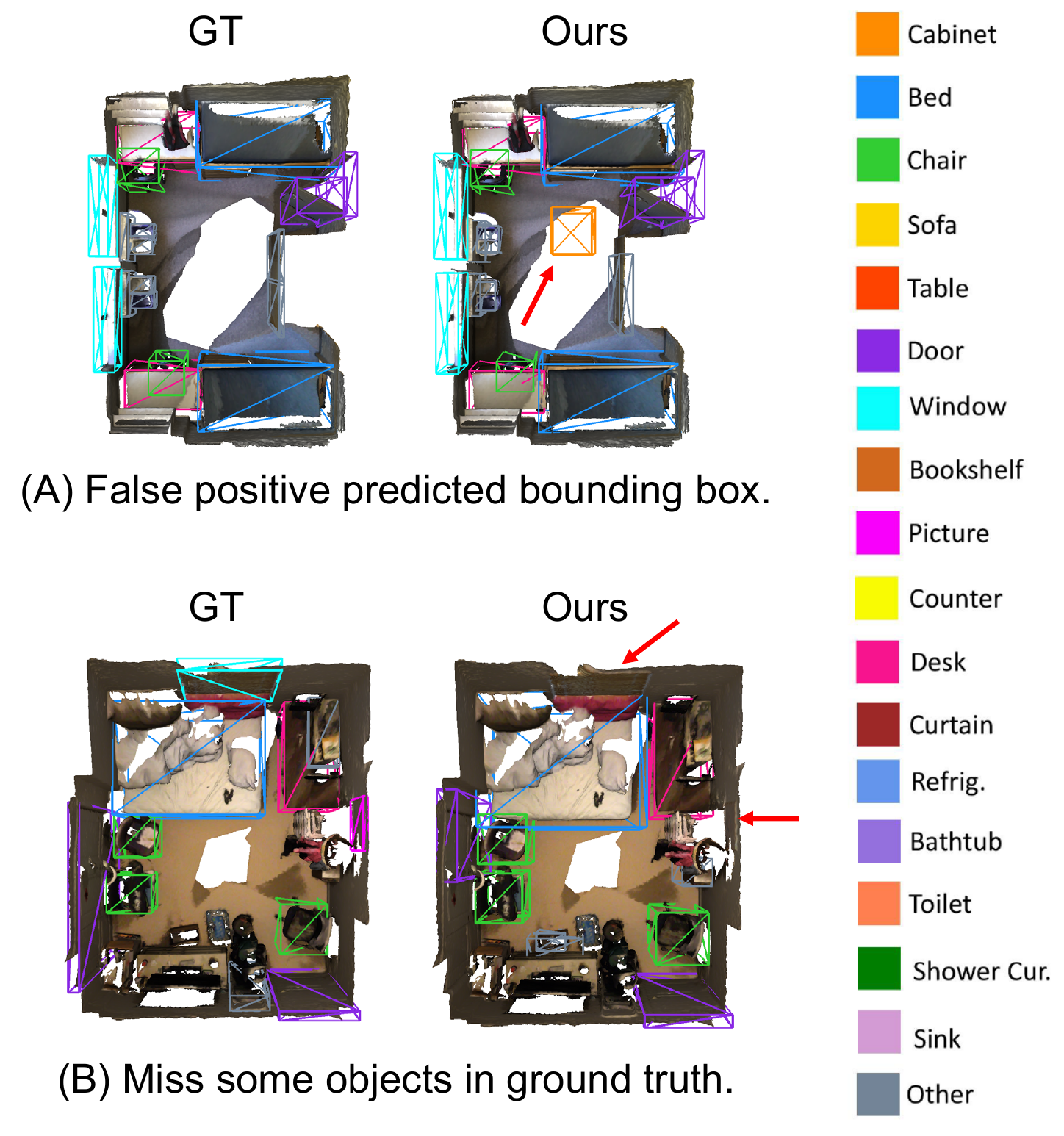}
    \caption{Samples of failure cases on ScanNet V2 dataset.}
\label{fig:failure-case}
\end{figure}


\begin{figure*}[t]
    \centering
    \includegraphics[width=\linewidth]{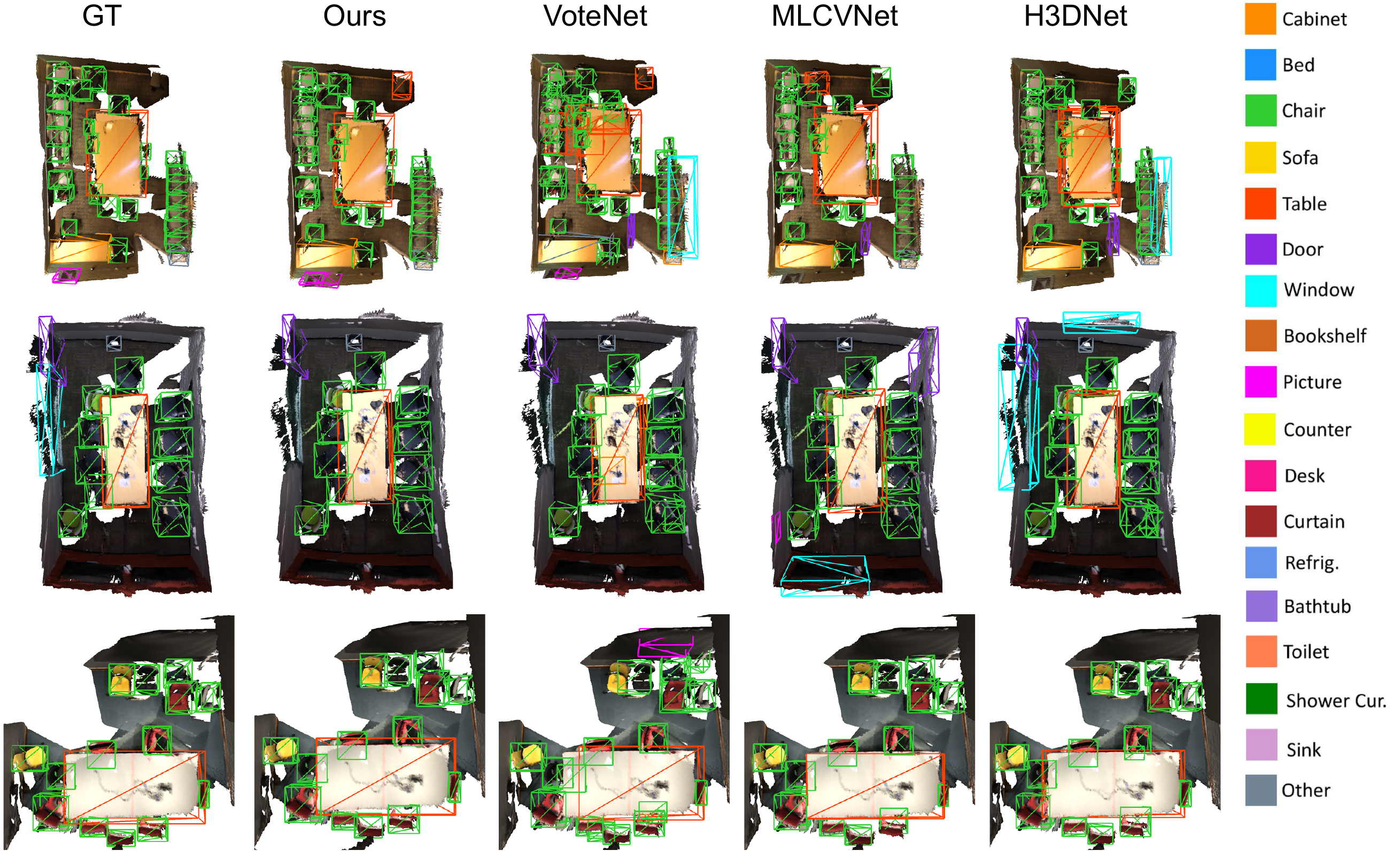}
    \caption{
    More qualitative results on ScanNet V2 dataset~\cite{scannet}.
    The reference methods are VoteNet~\cite{votenet}, MLCVNet~\cite{mlcvnet} and H3DNet~\cite{h3dnet}. Best viewed on screen.
    }
\label{fig:supple-scannet-viz}
\end{figure*}

\begin{figure*}[t]
    \includegraphics[width=\linewidth]{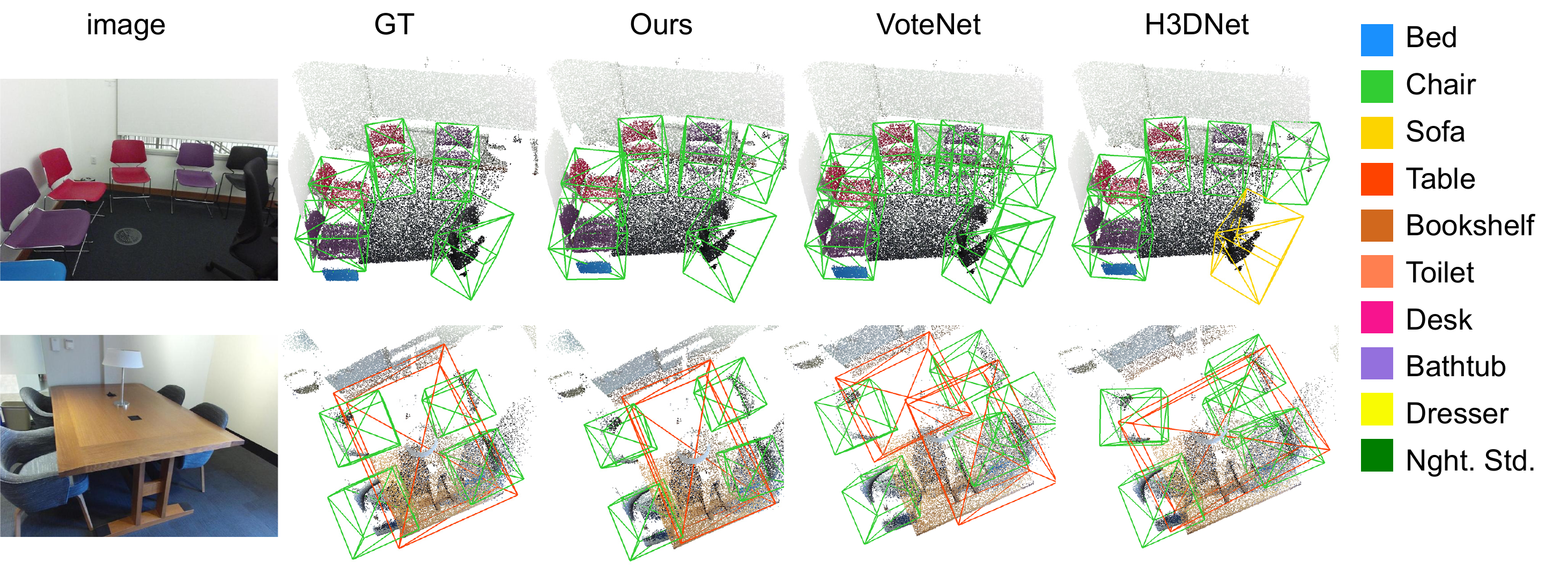}
    \caption{
    More qualitative results on SUN RGB-D dataset~\cite{sunrgb-d}.
    The reference methods are VoteNet~\cite{votenet} and H3DNet~\cite{h3dnet}. Best viewed on screen.
    }
\label{fig:supple-sunrgbd-viz}
\end{figure*}

\section{Supplementary}
This supplementary provides more quantitative results of our method (Sec.~\ref{subsec:quantitative-results}), more qualitative results (Sec. ~\ref{subsec:quantitative-results}), and finally implementation details (Sec. ~\ref{subsec:implementation-details}).

\subsection{More Quantitative Results}
\label{subsec:quantitative-results}

\noindent\textbf{Finer performance evaluations.}
We try to evaluate our method using mean average precision with multiple IoU thresholds for finer performance evaluations in Table~\ref{scannet@0.75} and~\ref{sunrgbd@0.75}.
We use mAP@0.25, mAP@0.50, mAP@0.75 to evaluate different methods, \ie VoteNet~\cite{votenet}, HGNet~\cite{hgnet}, MLCVNet~\cite{mlcvnet}, H3DNet~\cite{h3dnet} and our \method.
Our method performs the best on the metrics mAP@0.50 and mAP@0.75. Notably, mAP@0.75 requires more than $90\%$ coverage in each dimension of a bounding box, which is very challenging for a detector.
Our method gains $-1.1$\%, $2.8$\%, $3.7$\% increase on mAP@0.25, mAP@0.50, mAP@0.75 compared with H3DNet~\cite{h3dnet} using $4$ PointNet++ backbones and doubled input point clouds (\ie, $20,000$ points by our \method, and $40,000$ points by H3DNet)  on the ScanNetV2 dataset.
On more challenging evaluation metrics, our method has more gain, which shows the importance of our representative point generation, and its benefits for seed points revisiting and finer surface feature extraction to accurately detect objects with more reliable bounding boxes.

\noindent\textbf{Per-category results.}
We show the per-category results on ScanNet V2 dataset with 3D IoU threshold 0.25 in Table~\ref{scannet@0.25}, and the per-category results on SUN RGB-D with both 3D IoU thresholds 0.25 and 0.50 in Table~\ref{sunrgbd@0.25} and~\ref{sunrgbd@0.50}.
In terms of the accuracy about the object detection, our approach outperforms the baseline VoteNet~\cite{votenet} and prior state-of-the-art method H3DNet~\cite{h3dnet} significantly. 
For objects in the SUN RGB-D dataset, our approach can gain $7.9$\%, $10.9$\%, $6.7$\%, $7.6$\%, $6.3$\% increase on Bathtub, Bed, Dresser, Nightstand and Sofa compared with H3DNet~\cite{h3dnet}.
These improvements are achieved by using back-tracing and seed points revisiting to better capture object surface features.

\subsection{More Qualitative Results}
\label{subsec:qualitative-results}

We provide more qualitative comparisons between our method and the top-performing reference methods, such as VoteNet~\cite{votenet}, MLCVNet~\cite{mlcvnet} and H3DNet~\cite{h3dnet}, on the ScanNet V2 and SUN RGB-D datasets, as shown in Fig.~\ref{fig:supple-scannet-viz} and Fig. \ref{fig:supple-sunrgbd-viz}, respectively.
Our method can generate high-quality and compact predicted bounding boxes compared with the other reference methods.
We also show two typical failure cases in Fig.~\ref{fig:failure-case}. Our \method cannot avoid the existence of false positive predicted bounding boxes which appear on the hollow floor. Also, it is hard for our method to detect objects on the smooth wall, especially windows and pictures.
We need to mention that these failure cases are also common, and hard for the reference methods.
It is an interesting and significant future direction of our work to tackle these false positives when points are over sparse and increase the robustness when perceiving objects within the cluttered background. 

\subsection{Implementation Details}
\label{subsec:implementation-details}

As mentioned in the main paper, the \method consists of four modules: 
(1) vote generation and clustering, 
(2) back-traced representative points generation, 
(3) seed points revisiting, and 
(4) proposal refinement and classification followed by 3D NMS. 
Here we elaborate the implementation details with respect to each module.

\vspace{+1mm}
\noindent\textbf{Vote generation and clustering.}
We follow the same network architecture and vote regression loss as in VoteNet~\cite{votenet}.

\vspace{+1mm}
\noindent\textbf{Representative point generation.}
It has output sizes of $128$, $128$, $6+2\times NH$ for the three MLP layers, where $NH$ is the number of heading bins for estimating the orientations, $6$ is the $6$ distance offsets from vote point to object surface (front/back/left/right/up/down) in the canonical coordinate centered at the vote point. Then we sample $2$ representative points on each skewed direction as the back-traced representative points, thus we have $12$ representative points per proposal.

\vspace{+1mm}
\noindent\textbf{Seed point revisiting.}
We use the set abstraction module (SA module) to aggregate seed points features within $0.2$m radius surrounding a back-traced representative point. The SA module has the output size of $128$, $64$, $32$ for the MLP layers. After revisiting seed points, we get a $32$ dimensional feature vector for each representative point. We concatenate the representative point features in a predefined local-structure-aware order to a $32\times12=384$ dimensional feature vector per proposal. The feature vector is then projected to $128$-dimensional as the captured surface feature of the object proposal.

\vspace{+1mm}
\noindent\textbf{Proposal refinement and classification.}
The input is the $256$ dimensional fused feature vector which is the concatenation of $128$-D vote cluster feature and $128$-D revisited seed point feature.
Then the fused feature is fed into a  three-layer MLP, whose output sizes are $128$, $128$, $9+N_C$. $N_C$ is the number of semantic classes, \ie, $N_C = 10$ for the SUN RGB-D dataset~\cite{sunrgb-d} and $N_C = 18$ for ScanNet V2 dataset~\cite{scannet}.
In the first 9 channels, the first two are for objectness classification, the following one is for heading angle refinement and the last six are for distance offsets refinement.

\end{document}